# Interpreting deep learning output for out-of-distribution detection

Damian Matuszewski[1], Ida-Maria Sintorn[1,2]
[1]Department of Information Technology, Uppsala University, Uppsala, Sweden
[2]Vironova AB, Gävlegatan 22, Stockholm, Sweden
damian.matuszewski@it.uu.se, ida.sintorn@it.uu.se

## Abstract

Commonly used AI networks are very self-confident in their predictions, even when the evidence for a certain decision is dubious. The investigation of a deep learning model output is pivotal for understanding its decision processes and assessing its capabilities and limitations. By analyzing the distributions of raw network output vectors, it can be observed that each class has its own decision boundary and, thus, the same raw output value has different support for different classes. Inspired by this fact, we have developed a new method for out-of-distribution detection. The method offers an explanatory step beyond simple thresholding of the softmax output towards understanding and interpretation of the model learning process and its output. Instead of assigning the class label of the highest logit to each new sample presented to the network, it takes the distributions over all classes into consideration. A probability score interpreter (PSI) is created based on the joint logit values in relation to their respective correct vs wrong class distributions. The PSI suggests whether the sample is likely to belong to a specific class, whether the network is unsure, or whether the sample is likely an outlier or unknown type for the network. The simple PSI has the benefit of being applicable on already trained networks. The distributions for correct vs wrong class for each output node are established by simply running the training examples through the trained network. We demonstrate our OOD detection method on a challenging transmission electron microscopy virus image dataset. We simulate a real-world application in which images of virus types unknown to a trained virus classifier, yet acquired with the same procedures and instruments, constitute the OOD samples.

## Introduction

Deep learning (DL) shows great promises in various biomedical and microscopy image analysis applications as shown e.g., in these recent reviews on deep learning in medical image analysis [1], in medical and biomedical (pathology) applications [2], in image (microscopy) based cell analysis [3], and in different aspects of electron microscopy, also including biomedical applications [4]. However, for these methods to achieve their full potential and gain acceptance in a clinical and diagnostic setting, the knowledge and understanding of how and when they can be used and trusted need to be increased.

Employing DL methods in computer-aided diagnosis (CAD) requires that they are trusted by the medical staff, governing organizations, and by the general public. To achieve this, we must improve our understanding of their training, decision-making, and, most of all, their limitations. One way to increase credibility is to admit where knowledge lacks. Therefore, a method that would signal that a model is unsure of its decision or that such a decision is clearly beyond its current capabilities is of the highest interest.

In machine learning, we often describe problems as either supervised or unsupervised. However, in many practical scenarios, the classification problem is something in between. Classifiers tend to fail when employed in real-world tasks and exposed to out-of-distribution samples (OODs), i.e., instances unlike those in the training set [5, 6]. In the literature, they are sometimes also referred to as outliers and anomalies if they are rare unusual examples of the training set classes, or novelties if they are from entirely new, untrained-for classes. The training set is always limited to the acquired samples and their possible augmentation. Yet in many real applications such as ambient monitoring, medical diagnosis, virology, cytometry, drug discovery, etc., an algorithm running in the "wild" will have to face many instances of rare or unexpected classes that were not included in its training set. These samples are silently "forced" to be classified as one of the known classes, often with high confidence [5, 6]. For example, in virology, a classifier may consistently classify with high confidence new virus species or subspecies, even though it should flag them as difficult examples for human intervention because recognizing their true class is outside its capability. This silent failing to recognize OODs or to admit lack of competence can substantially limit Artificial Intelligence (AI) adoption and deployment in many applications, cause serious accidents, and greatly reduce the trust in AI-based diagnosis.

Here, we investigate a DL model output to better understand its behavior when exposed to OOD samples. We analyze the distributions of the raw model outputs (logits) of the training set and two independent datasets unused during the model development: (1) test set with instances of classes known to the model, and (2) OOD set with instances of classes outside the competence of the model. Based on our observations, we developed a novel approach that relaxes supervised DL allowing semi-supervised inference. Our method detects rare albeit important outliers in supervised classification, i.e., in models trained to assign new samples to one of a set of known classes. It also provides probabilities for an input sample belonging to each of the known classes. It is backward compatible with already trained and applied Artificial Neural Network (ANN) classifiers, i.e., it does not require any architecture modification, it is possible to apply without the necessity of resource-consuming model retraining or specific OOD examples for fine-tuning. All that is required are the inference of the training set to establish the data distributions and the corresponding per-class probabilities, and replacing the final activation function (softmax) with our OOD sample detector.

It is important to evaluate OOD detection methods on realistic data that comes from similar instruments and is captured with the same intention. Most of the published approaches use one natural scenes dataset for training a classifier and a completely different image dataset as the OOD examples [7, 8]. This is equivalent to e.g., using microscopy images for training a model and using CT scans as OOD examples. Such situations are very unlikely, and thus, such evaluations simulate poorly real-world applications. In addition to the commonly used reference datasets Cifar10 and Cifar100 [9] which share many characteristics and are considered more realistic for OOD detection scenarios, we demonstrate and evaluate the approach on virus classification in transmission electron microscopy (TEM) images, representing a plausible real-world biomedical application example.

## Background

A typical supervised ANN classifier has one output (i.e., one neuron in the last layer) for each of the possible classes. Softmax is the most commonly used activation function in the last layer. It scales and normalizes the raw neuron outputs $x$ so that the sum of all final output values in that layer (typically the network prediction results) is equal to 1:

$$softmax(x_n) = \frac{e^{x_n}}{\sum_i e^{x_i}}. \tag{1}$$

For a given input sample and an ANN classifier, the softmax output vector values are often interpreted as the network confidence of that sample's affiliation to corresponding classes. While softmax is very popular thanks to its easy implementation and interpretation, it is not free of problems. First, as softmax imposes that the prediction results sum up to 1, any sample that is an outlier belonging to none of the expected classes will be forced to be classified as one of the known classes. This, even if the raw output values from all neurons are close to zero or negative which could be interpreted as the absence of the input signal combination that the neurons were trained to detect. Indeed, Hendrycks and Gimpel have demonstrated that random Gaussian noise fed into an MNIST image classifier gives "a positive prediction confidence" of 91% [7]. Second, the exponential normalization means that softmax may often hide the confusion or uncertainty of the model. Consider the following example: suppose there are 4 classes and the 4 raw network outputs are {6; 5; 0; 1}, the corresponding softmax output is {0.726; 0.267; 0.002; 0.005}. Now, suppose the raw network output is {1,000,006; 1,000,005; 0; 1} the corresponding softmax output is {0.731; 0.269; 0; 0}. In both cases, softmax gives the highest score by a large margin to the first class, even though the raw network output is similarly high for the second class. This occludes the confusion and lack of confidence of the model; in these examples, the output "probability" of the second class should be comparable with that of the first class. Finally, if the number of classes is large (hundreds or thousands) the denominator can become large and softmax tends to give low scores distributed over many classes.

## Related work

Several different approaches for handling the softmax function shortcomings and OOD detection have been suggested. They can be roughly categorized as either using OODs for parameter tuning or not and/or either modifying the network architecture which requires retraining or not.

Hendrycks and Gimpel [7] proposed a simple approach based on the observation that correctly classified examples tend to have greater maximum softmax probabilities than erroneously classified examples and OODs. They deployed simple thresholding to the maximum softmax results for detecting OODs in new samples. Liu et al. [10] used an energy function of logits instead of softmax to detect OOD samples. Their method can be applied either directly to pretrained models or used in model fine-tuning. Both of these approaches are similar to our method as neither requires modifications of the network architecture, retraining, nor access to OOD examples. Moreover, all three rely on the discriminant power of the model to correctly classify in-distribution samples. Other methods mentioned in this section may achieve better performance, however, they require substantial changes to the architecture (often adding an additional network), retraining, and/or OOD samples for fine-tuning. As they cannot be easily applied to already trained and deployed models, they have different target applications. Therefore, we focused our method evaluation on comparing it to the Liu et. al and Hendrycks and Gimpel methods which we from here on refer to as the *Energy* and *H&G baseline methods*, respectively.

Liang et al. [11] suggested using softmax temperature scaling and adding small, controlled perturbations to the inputs, to improve the separation between the in-distribution (ID) and OOD samples. Their ODIN method uses OOD examples to tune the parameters and relies on the analysis of the softmax results, and hence, it is affected by softmax's aforementioned problems. In [12], a generalized ODIN is suggested that doesn't require access to OOD samples for hyperparameter tuning. However, their method requires modification of the network's last layers and retraining. The same holds for [13] in which they modify the network's output layer to include a module for deriving scaled cosine similarity on which softmax is applied.

Abdelzad et al. [14] used features from an intermediate layer of a DL classifier. They hypothesize that there is a latent space (feature layer) in which the distribution of OODs is well separated from IDs. Although it is described as a method that does not require OOD examples for training the detector, the OOD samples are used to choose the network layer for feature extraction, i.e. the latent space in which OODs are best separated from IDs. Ahuja et al. [15] and Lee et al. [16] model the class distributions on features in internal deep layers by different probabilistic models and use that as uncertainty estimates in the classification. Also, Sastry and Oore [17] use the output from intermediate layers of a deep model classifier in combination with the output class assignment to detect OOD samples via Gram Matrices comparison – an idea popularized in artistic style transferring with deep learning models.

Another approach to identify OODs is through dimensionality reduction and visualization and/or clustering of selected ANN features. Eulenberg et al. used the learned network features before the softmax classifier (i.e., the one-but-last fully connected layer) to visualize continuous biological processes (cell-cycle in Jurkat cells and disease progression in diabetic retinopathy) and to detect outliers in an unsupervised manner [18]. Similarly, Faust et al. performed the unsupervised classification and outlier detection in surgical neuropathology [19].

Yet another approach is to modify the network architecture and retrain to achieve simultaneous classification, anomaly detection, and estimation of the prediction confidence. That is suggested in e.g. [7 (as an alternative to the H&G baseline method), 8, 10, 20, 21]. These ANNs then detect both OODs and misclassifications by scoring each prediction.

Related to OOD detection is the well-studied problem of outlier detection, reviewed in [22]. Chen et al. [23] and Liu et al. [24] both propose methods for training an ANN dedicated to outlier detection. However, these techniques are implemented as a separate pre-processing tool rather than OOD / misclassification detection built-in to an ANN classifier. Sommer et al. acknowledge the supervised learning dependence on a priori knowledge of the possible classes [25]. They use a combination of deep learning and human-engineered features in anomaly detection, where new samples are compared to known distributions of negative controls.

## Context virus dataset

To describe and evaluate our approach, we use the Context Virus Dataset [26]. It contains image patches cropped from 1245 images of 22 virus classes captured with two different transmission electron microscopes. The 14 classes with the largest number of particles are used in training the DL classifier while the remaining 8 classes are left out to represent and illustrate new, unknown viruses – OOD samples. Table 1 summarizes the dataset. We use the proposed representative split into training, validation, and test sets from

[26], as well as the suggested augmentation of the training set. The augmentation (image flipping and multiple 90 degrees rotation) results in an equal number of image patches (736) for all classes used in training.

The virus image patches in the dataset are grayscale images of 256 x 256 pixels with a fixed pixel size of 1 nm. Although the target particle is always in the center of the patch, parts of other or even entire virus particles can also be present in the images (some particles in the dataset are single while others are clustered). However, these are always of the same type as the target particle. Before passing the images to the DL models we normalized them by subtracting the mean from each image patch and dividing by its standard deviation.

Table 1. TEM virus dataset. The virus classes used as OODs in testing are in bold and marked with an asterisk.

| Virus | av. particle size [μm] | # image patches | | | |
|---|---|---|---|---|---|
| | | Train | val. | test | Total |
| Adenovirus | 80 | 736 | 44 | 86 | 866 |
| Astrovirus | 25 | 736 | 86 | 66 | 888 |
| CCHF | 120 | 736 | 100 | 86 | 922 |
| Cowpox | 270 | 736 | 72 | 59 | 867 |
| *Dengue | 45 | - | - | 131 | 131 |
| Ebola | 80 | 736 | 469 | 363 | 1568 |
| *Guanarito | 140 | - | - | 48 | 48 |
| Influenza | 110 | 736 | 281 | 170 | 1187 |
| Lassa | 140 | 736 | 114 | 128 | 978 |
| *LCM | 120 | - | - | 55 | 55 |
| *Machupo | 120 | - | - | 108 | 108 |
| Marburg | 80 | 736 | 858 | 858 | 2452 |
| Nipah | 95 | 736 | 40 | 35 | 811 |
| Norovirus | 30 | 736 | 104 | 84 | 924 |
| Orf | 145 | 736 | 76 | 31 | 843 |
| Papilloma | 55 | 736 | 227 | 187 | 1150 |
| *Pseudocowpox | 145 | - | - | 85 | 85 |
| Rift Valley | 90 | 736 | 414 | 392 | 1542 |
| Rotavirus | 80 | 736 | 48 | 40 | 824 |
| *Sapovirus | 30 | - | - | 38 | 38 |
| *TBE | 50 | - | - | 62 | 62 |
| *WestNile | 50 | - | - | 188 | 188 |
| Total OOD | | | | | 715 |
| Total | | 10304 | 2933 | 3300 | 16537 |

# Convolutional neural network architecture

Our OOD detection strongly relies on the discriminant power (the ability to correctly identify in-distribution samples) of the classifier underneath. Therefore, following the virus classification comparison results reported in [26], we chose DenseNet201 [27] pre-trained on ImageNet [28] to describe our raw output analysis and the OOD detection approach. The framework architecture for model fine-tuning is presented in Fig 1. In this case, we replaced its original last layers with 3 fully connected layers with interposed dropouts. We fine-tuned the pre-trained model for 50 epochs and used the categorical cross-entropy loss function, the

Adam [29] optimizer, and early stopping based on the validation performance. The model achieved the best performance after around 40 epochs. The model was implemented using TensorFlow [30] and Keras [31]. Our code, preprocessed data, and model weights are available at [32]. Raw virus images are available at [33].

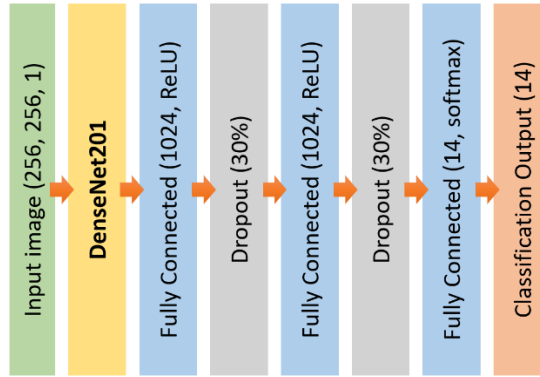

**Fig 1. The transfer learning framework for DenseNet201.**

# Raw network output analysis

Typical DL classifiers have dense (fully-connected) layers performing the final task of assigning a sample to one of the predefined classes. In such cases, the last layer is typically followed by softmax as its activation function and the number of neurons in this layer corresponds to the number of predefined classes. Each neuron in the last layer is connected to one output and can be interpreted as a stand-alone detector of that class. It takes the output of the previous layer (the ultimate feature vector produced by the model) and based on this evidence it assesses whether the sample belongs to its specific class. DL model training ensures that the presence of specific target patterns is highlighted and propagated through the network as high positive values and their absence – by low or negative. This is also true for the neurons in the last, fully connected layer. Each neuron returns high values if the evidence (input) it received indicates its corresponding class or low (close to zero or negative) if the evidence is against the sample belonging to the node's class. This information is lost when softmax is used for interpreting and normalizing the outputs of all nodes in the last fully connected layer.

Fig 2 shows the distribution of raw network output (logits) for the neuron assigned to one (Norovirus) class just before the last activation function (softmax) and the three data subsets: training, validation, and test. We can observe that these values form two separate distributions: one for the samples of the correct class and one for all other samples. Each neuron works as a pattern detector that returns high values if its dedicated pattern is present in the input signal and low – in all other cases. In the case of the last layer, a neuron simply decides whether the features extracted from the input give enough evidence (both in strength and combination) to be classified as the corresponding class. The stronger the evidence, the higher the output; and contrary: the more evidence against the dedicated class for that neuron the lower the output; thus, creating the two separate distributions.

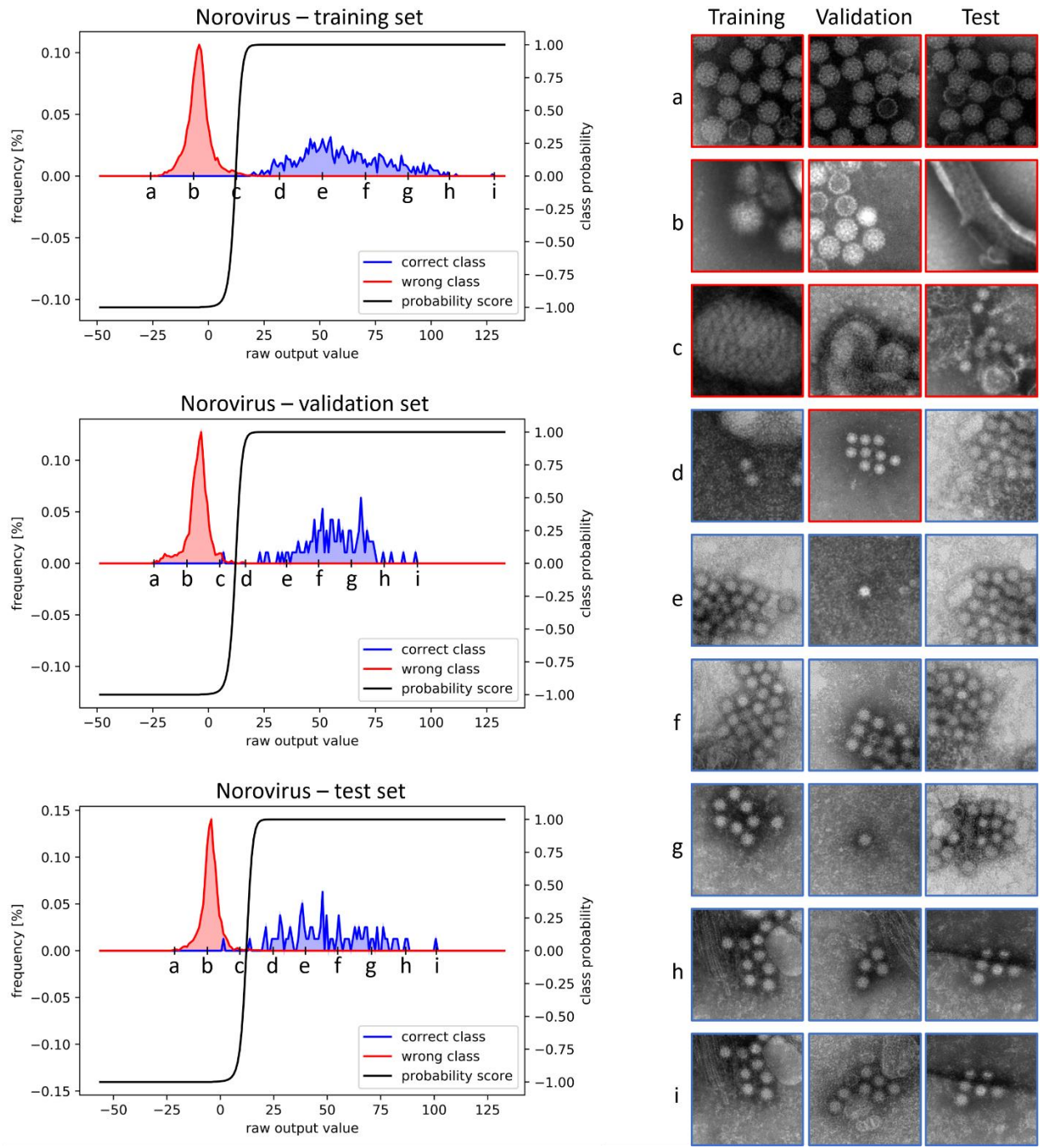

**Fig 2. Output logit distribution.** Left: distributions of the raw network output for the neuron responsible for detecting Norovirus and the three data subsets: training, validation, and test. Right: sample images from the dataset with specific raw network output values. The corresponding positions (raw output values) of the images are shown in the plots.

The distributions are sometimes better separated in the training set. In some cases, as exemplified in Fig 3, the distributions begin to overlap in the validation and/or test sets. This may be caused by the training set not being representative enough. The neurons are confused by the new samples and their output for the correct classes is lowered. If there is enough training data to represent the full variation of the class in the test set, the per-class raw values show a clear separation between the correct class for that neuron and samples from all other classes, as shown in Fig 2.

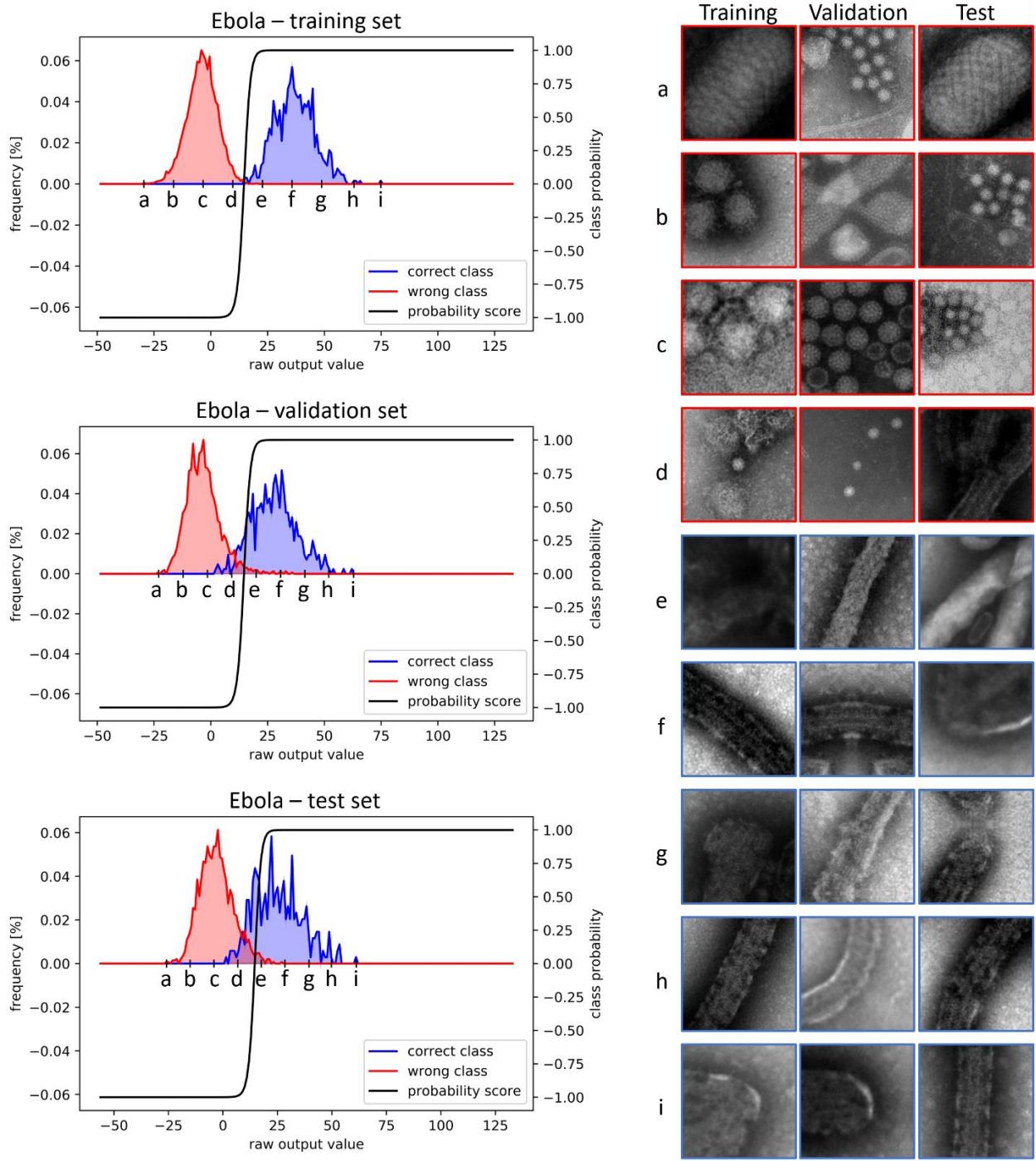

**Fig 3. Output logit distribution.** Left: distributions of the raw network output for the neuron responsible for detecting Ebola and the three data subsets: training, validation, and test. Right: sample images from the dataset with specific raw network output values. The corresponding positions (raw output values) of the images are shown in the plots.

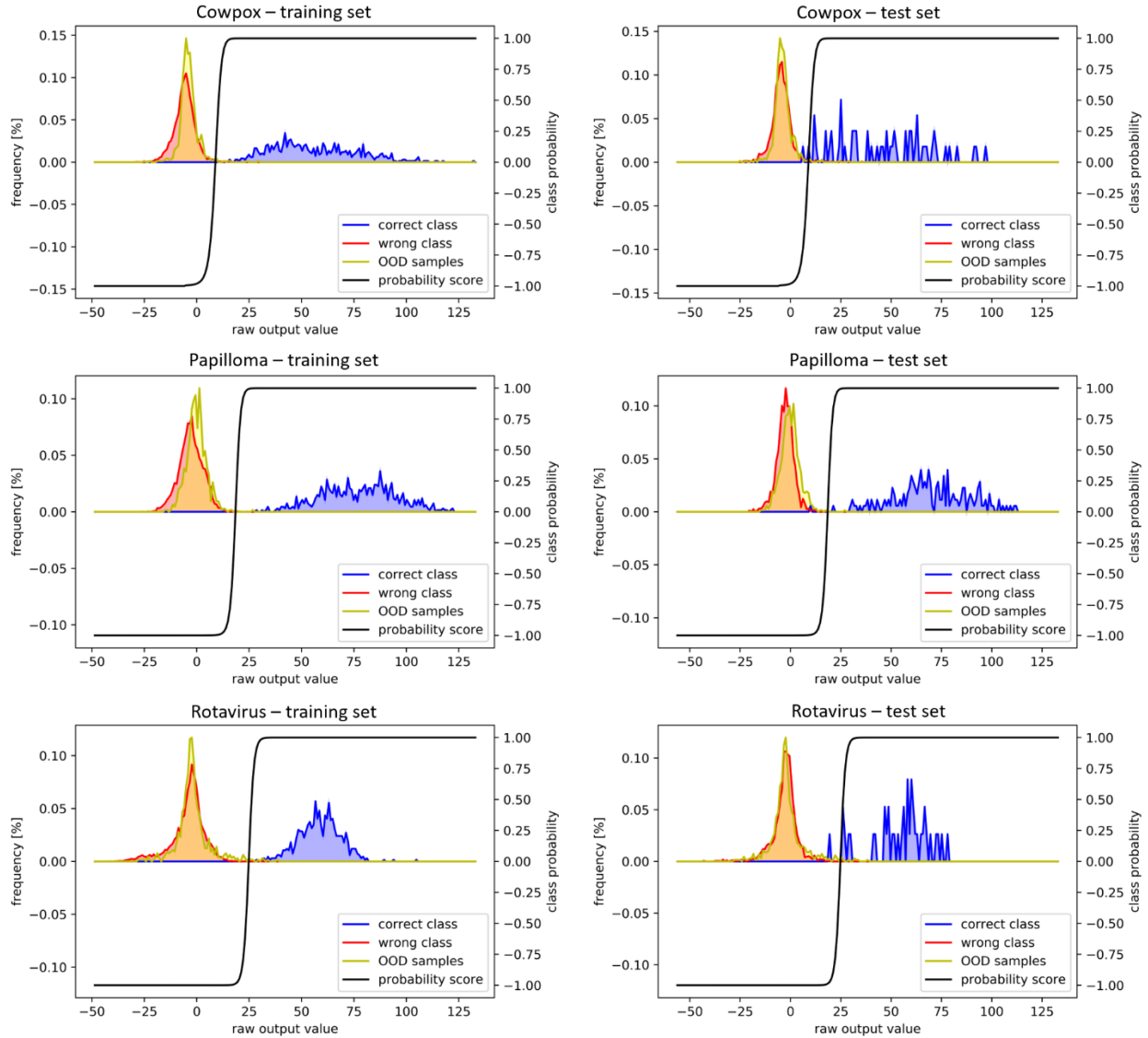

**Fig 4. Distributions of the raw network outputs for the neurons responsible for detecting Cowpox, Papilloma, and Rotavirus, respectively from top to bottom. The left column presents training set samples and the right column test set samples. The OOD distributions are from the same OOD dataset in all subfigures.**

Fig 4 shows the distributions of the raw network outputs for the OOD dataset overlayed on the training and test sets for neurons responsible for Cowpox, Papilloma, and Rotavirus recognition. We can observe that the OOD distributions (yellow) overlap well with the distributions of the samples not belonging to the respective neuron class (red). This observation inspired us to develop an OOD detector based on the distributions of raw network outputs described in the next section. The corresponding plots for all virus classes in the dataset are shown in the Supporting Information (S1 File).

The neurons in the last layer are independent of each other: they do not share weights and in the backpropagation training they receive their own weight updates. As a consequence, their output naturally is different – tuned to detect specific distinct patterns. Moreover, their output distributions for the training set samples differ as well, as can be observed in Fig 4. Both correct and wrong class distributions have different characteristics between the neurons (and their corresponding classes). Especially the correct class distributions vary substantially both in mean (ranging from 37.6 to 77.7) and in standard deviation (ranging from 8.8 to 20.4). This means that a specific raw output value does not have the same class support (or meaning) in one neuron as in another. Therefore, the raw network output should not be compared (e.g., with softmax) without normalization for the differences in the expected distributions. Table 2 presents the mean and standard deviations of each class's raw network output values in the training set for the corresponding output neuron.

**Table 2. Mean and standard deviation of the correct raw network output values in the training set for the corresponding output neuron.**

| Virus class | Mean | Std dev |
|---|---|---|
| Adenovirus | 54.6 | 11.4 |
| Astrovirus | 64.2 | 17.0 |
| CCHF | 49.4 | 12.4 |
| Cowpox | 56.3 | 20.4 |
| Ebola | 37.6 | 8.8 |
| Influenza | 56.5 | 14.5 |
| Lassa | 38.3 | 9.9 |
| Marburg | 38.3 | 8.8 |
| Nipah | 37.8 | 11.4 |
| Norovirus | 60.2 | 19.1 |
| Orf | 76.6 | 17.4 |
| Papilloma | 77.7 | 17.8 |
| Rift Valley | 44.0 | 9.0 |
| Rotavirus | 59.2 | 9.7 |

# Out-of-distribution detection

## Per-class probability scores

We use the raw network outputs of the training set to compute the per-class probability scores (PS). With the assumption that the distributions are normal, we trained a separate Gaussian Naïve Bayes (NB) classifier for each class to obtain the probability estimates for a new sample belonging to one of the two distributions (class or outlier). These NB classifiers can be understood as a set of one-class classifiers, each with only one input – the raw output of the corresponding neuron in the last layer of the model. Since the distributions are strongly unbalanced (samples from a single class versus all others) we reset the a priori probabilities to 0.5. Next, we subtracted the NB classifier's "wrong" class probability estimate from the "correct" class probability estimate and set the resulting PS to -1 and 1 for values smaller than the mean of the wrong class distribution and bigger than the mean of the correct class distribution, respectively. The resulting function is bound to [-1,1] and takes negative values for samples from the wrong class distribution and positive values – for the samples from the correct class distribution, see Figs 2-4. The higher the absolute PS value the higher the probability (the classifier's confidence that the sample belongs to that specific distribution). Once the PS functions are computed for the training set, they can be applied at inference as a clue to OOD detection. In all illustrative Figs for the validation, test, and OOD sets, the PS functions (black curves) were calculated using the corresponding training set.

As PS functions are computed for each class separately, they are adapted for the different neuron output distributions. They have different slopes and decision boundaries that follow the two distributions for each neuron, as shown in Figs 2-4.

## Probability scores interpreter

While individual per-class PSs are useful for estimating the confidence of a model, interpretation of their values for all classes as a set can be used for detecting OOD samples. For that purpose, we designed the probability scores interpreter (PSI) summarized in Table 3. It results in one of the three codes: green, yellow, and red. The green code means that the model is confident about its prediction and the class with the highest positive PS can be taken as the final classification for that sample. It happens when only one per-class PS is positive and its value is high. The yellow code signalizes that the confidence of the model is lowered by potential confusion with another, similar class. It happens when several classes have positive PS but only one of them has a high value. The red code indicates that the model can not classify unambiguously the sample and that it requires a manual inspection. It happens when there are no positive PSs or there are many positive per-class PSs – either with high values or there is no single one with a high positive value. These cases may indicate an OOD sample (e.g., an unknown class or imaging malfunction) or data variation that was not represented in the training set.

Table 3. The global interpretation of the per-class probability scores.

| Positive values | Negative values (all remaining) | Interpretation | Code |
| --- | --- | --- | --- |
| 1 high | High | Clear result | Green |
| 1 high | High, 1+ low | Border case | Green |
| 1 high, 1+ low | High or low | Border case | Yellow |
| 2+ high, 0+ low | High or low | Confusing evidence, difficult sample, potentially OOD | Red |
| 1+ low | High or low | Not enough evidence, difficult sample, potentially OOD | Red |
| None | All high or low | OOD sample | Red |

Fig 5 presents examples from the test and OOD sets that fall into different categories from the PSI look-up table. It also shows the corresponding PSI category, maximum softmax results, true label, and the predictions obtained with PSI, softmax, and the H&G baseline method [7]. We can observe that even though the threshold in the H&G baseline method was set to the optimal $1 - 10^{-6}$, it still missed the three difficult OOD samples (4-6 in Fig 5). In fact, the threshold was so high that the H&G baseline method labeled the three test samples with high maximum softmax output as OODs (1-3 in Fig 5). These examples demonstrate the shortcomings of softmax and methods that are based on it.

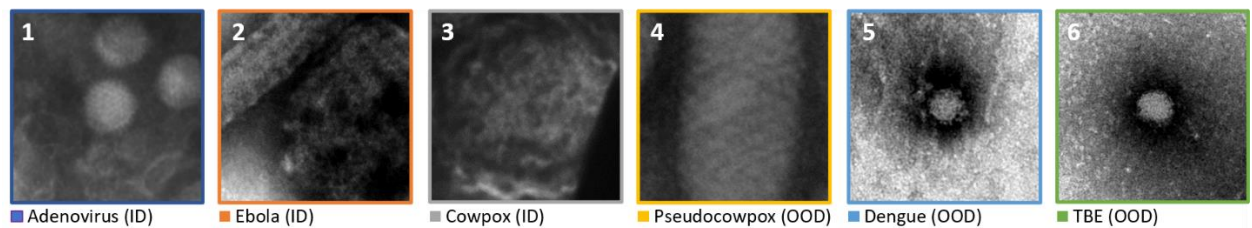

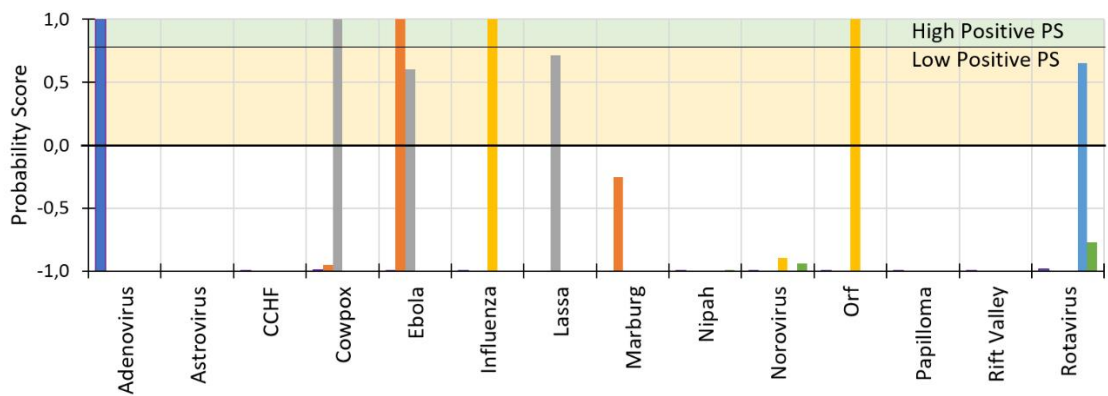

| Sample | Probability Score Interpreter | | | Classification | | | | True Label |
|---|---|---|---|---|---|---|---|---|
| | Positive values | Negative values | Interpretation | PSI | Baseline | Max softmax | | |
| 1 | 1 high | High | Clear result | Adenovirus | OOD | 0.999863 | Adenovirus | Adenovirus |
| 2 | 1 high | High, 1+ low | Border case | Ebola | OOD | 0.999996 | Ebola | Ebola |
| 3 | 1 high, 1+ low | High or low | Border case | Cowpox | OOD | 0.982145 | Cowpox | Cowpox |
| 4 | 2+ high, 0+ low | High or low | Confusing evidence, difficult sample, potentially OOD | OOD | Orf | 1.000000 | Orf | OOD (Pseudocowpox) |
| 5 | 1+ low | High or low | Not enough evidence, difficult sample, potentially OOD | OOD | Rotavirus | 1.000000 | Rotavirus | OOD (Dengue) |
| 6 | None | All high or low | OOD sample | OOD | Rotavirus | 1.000000 | Rotavirus | OOD (TBE) |

**Fig 5. Examples from the test and OOD sets that fall into different categories from the PSI look-up table.** Top: examples from the test and OOD sets that fall into different categories from the Probability Score Interpreter look-up table. Middle: corresponding Probability Score values. Bottom: corresponding OOD detection and classification results. The predictions were obtained with the Probability Score Interpreter and the H&G baseline method [7] with the optimal thresholds for the virus dataset, i.e., $1 - 10^{-3}$ and $1 - 10^{-6}$, respectively.

# Implementation in existing models

The proposed OOD detection approach can be easily applied to already deployed networks with softmax as the last activation function. The only required work is running the inference on the training (and potentially validation) set to establish the distributions of raw output values for each class vs the rest of the dataset and from them compute the per-class probability scores. Once this is done, the softmax is replaced with the per-class probability scores and their interpreter. Table 4 summarizes the implementation steps in already trained and deployed models.

**Table 4. Implementation steps of our OOD method in deployed deep learning classifiers.**

| |
|---|
| 1. Remove softmax as the classifying activation function. |
| 2. Forward-pass the training set to get the distributions of raw output and compute the probability score functions for each class. |
| 3. Use the validation set to automatically select the threshold for the Probability Score Interpreter. |
| 4. Place the tuned interpreter at the classifying head. |

## PSI threshold effect

PSI uses a threshold $t$ to define what is considered an in-class sample for a given class. The PSI threshold is constant for all classes, however, as the probability scores are computed separately for each class, the PSI threshold corresponds to different logit values in each class. The number of classes that comply with the threshold is then interpreted according to the PSI look-up table (Table 3). Fig 6 demonstrates how the choice of the threshold affects the performance. As expected, the higher the threshold the more samples are being classified as OOD samples (outliers). This may lead to lower classification errors (by labeling problematic samples as OOD rather than classifying them erroneously) but it may also cause a higher false-positive OOD detection rate. Both very high or very low PSI threshold values give similar results: treating almost everything as OODs. That is because the OOD detection is based on a binary PS thresholding of each class separately and investigating the number of those that passed the threshold. A very high threshold means that only the "purest" samples are considered IDs and all others – OODs. A very low threshold means that many samples pass the threshold in more than one class which means that they are marked as OODs.

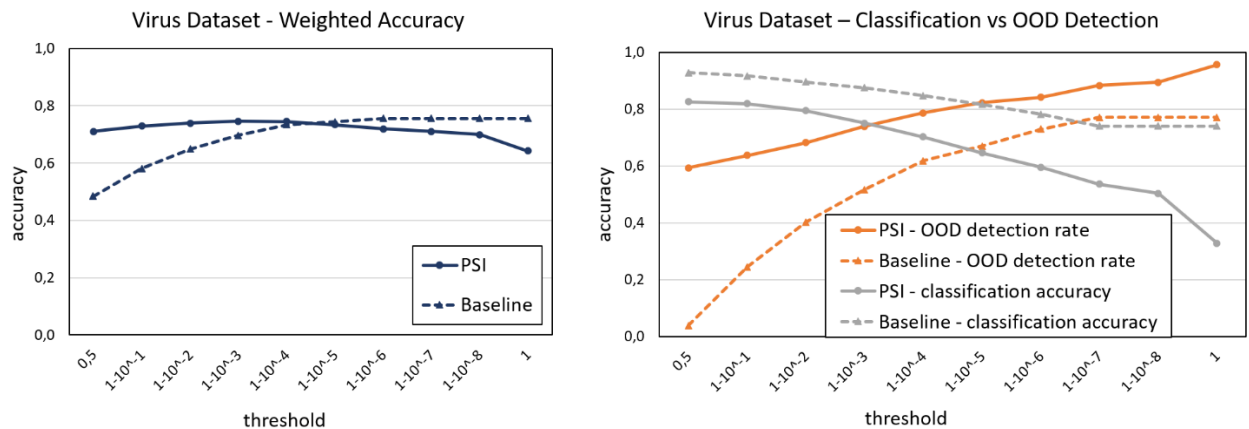

**Fig 6. Threshold effect on the performance of the PSI and the H&G baseline method.**

## Evaluation

**Performance Metrics**

Most of the current OOD detectors return a confidence score that can be thresholded to classify samples as either in-distribution (ID) or OOD [22]. By changing the threshold value, the user can vary the sensitivity of the OOD detector. Different threshold values are also used to generate Receiver Operating Characteristic (ROC) and Precision-Recall (PR) curves that are used as the basis for the performance evaluation of state-of-the-art OOD detectors in various scenarios and applications. However, the PSI threshold has a different meaning. It interprets the PS for each class, rather than the global confidence score, and thus, it cannot be used in the same way as the threshold in other OOD detectors to generate the ROC and PR curves (as explained in 5.4, both too high and too low PSI threshold values result in treating more samples as OODs). Therefore, instead of these metrics, we use the weighted accuracy when comparing PSI to the H&G baseline method. We defined the weighted accuracy as the arithmetic mean of the classification accuracy and the OOD detection rate:

$$Weighted\ Accuracy = \frac{n_T/N_T + n_{OOD}/N_{OOD}}{2}, \qquad (2)$$

where, $n_T$ and $n_{OOD}$ are the numbers of correctly classified test and OOD samples, and $N_T$ and $N_{OOD}$ are the total numbers of the test and OOD samples, respectively.

**Comparison to other methods**

Most of the OOD detectors have been evaluated on far-out-of-distribution datasets (i.e., OOD datasets that are very different from the ID training samples, e.g., as introduced by Hendrycks and Gimpel [7] and expanded by Liang et al. [11]). However, Hendrycks et al. [21] pointed out that detecting near-distribution anomalies (i.e., samples that are related to or very similar to the ID training examples) is far more challenging than detecting far-OOD samples. In real-world applications, near-OOD samples are much more likely to happen and, thus, they are more relevant. Therefore, in addition to the virus dataset, we limited our evaluation to only the near-OOD scenarios of detecting Cifar10 vs Cifar100 and vice versa. Cifar10 and Cifar100 [9] are computer vision datasets of natural image scenes, subsampled from the Tiny Image Dataset. Cifar10 contains images from 10 classes, whereas Cifar100 has 2 levels of labels: coarse with 20 classes and fine with 100 classes. Therefore, Cifar100 can be used to train 2 different DL models on the same dataset: one for the coarse classification and one for the fine classification. This results in 3 evaluation combinations: 1) Cifar10 as ID vs Cifar100 as OOD, 2) Cifar100 trained with course labels as ID vs Cifar10 as OOD, and 3) Cifar100 trained with fine labels as ID vs Cifar10 as OOD.

Authors of the methods that we used in our evaluation: the H&G baseline [7] and the Energy [10], did not specify how to set the thresholds for the OOD detection. We selected the threshold for each method and each dataset based on the corresponding validation set. In the virus dataset, we selected the minimum threshold that detected at least 75% of the incorrect validation set classifications as the OOD examples. In the Cifar cases, we used the thresholds corresponding to at least 90% of the incorrect validation set classifications as the OOD examples. We initially aimed to use the thresholds corresponding to 90% in all cases but the H&G baseline method could not achieve this in the virus dataset (unless using the threshold equal to 1 which would result in a useless detector that assigns all input to OOD). This way of determining the threshold without OOD examples has proven to be a good approximation of the "best" threshold in PSI and H&G baseline, except for the PSI in the Cifar100-fine vs Cifar10 scenario, where the threshold was set too high. In these cases, the "best" threshold was selected from a set of thresholds given by $1 - 10^{-x}$, where $x \epsilon \{1, 2, 3, ..., 8\}$ based on the weighted accuracy criteria. The same threshold set was used to create the plots in Fig 7.

Table 5 shows the OOD detection performance comparison in various datasets. For PSI and H&G baseline we report results obtained with two thresholds: (1) obtained from the validation set analysis (without access to the OOD examples), and (2) the best performing threshold based on the weighted accuracy criteria (denoted "(best)" in the table). Our method showed better performance than the Energy method and a similar performance to the H&G baseline approach. Except for Cifar-10 (ID) vs Cifar-100 (OOD), PSI showed a better OOD Detection Rate than the H&G baseline. In the original Energy method paper [10], the authors report the superior performance of their approach over the H&G baseline and several other OOD detection methods. However, they focused their evaluations only using far OOD samples, i.e., OODs in their evaluation are taken from very different datasets from those in the training set. In real applications, the OODs often show high similarity to the expected in-distribution samples. Our results show that the Energy method struggles when facing near OOD examples.

Table 5. Performance of the PSI and H&G baseline methods in various OOD detection scenarios.

| Dataset | Method | Threshold | Weighted Accuracy | Classification Accuracy | OOD Detection Rate |
|---|---|---|---|---|---|
| Virus Dataset | PSI (our) | 0.999828 | 0.746 | 0.712 | 0.779 |
| | | (best) $1 - 10^{-3}$ | 0.746 | 0.751 | 0.740 |
| | H&G Baseline [7] | 0.999997 | 0.749 | 0.794 | 0.705 |
| | | (best) $1 - 10^{-6}$ | 0.756 | 0.782 | 0.730 |
| | Energy [10] | 23.01 | 0.737 | 0.773 | 0.702 |
| ID: Cifar-10 OOD: Cifar-100 | PSI (our) | 0.999990 | 0.806 | 0.708 | 0.904 |
| | | (best) $1 - 10^{-4}$ | 0.817 | 0.783 | 0.850 |
| | H&G Baseline [7] | 0.993654 | 0.811 | 0.818 | 0.805 |
| | | (best) $1 - 10^{-3}$ | 0.816 | 0.728 | 0.903 |
| | Energy [10] | 11.13 | 0.801 | 0.693 | 0.910 |
| ID: Cifar-100-coarse Cifar-10 (OOD) | PSI (our) | 0.999984 | 0.718 | 0.590 | 0.845 |
| | | (best) $1 - 10^{-5}$ | 0.718 | 0.578 | 0.857 |
| | H&G Baseline [7] | 0.977878 | 0.688 | 0.662 | 0.715 |
| | | (best) $1 - 10^{-2}$ | 0.696 | 0.616 | 0.777 |
| | Energy [10] | 12.37 | 0.692 | 0.447 | 0.938 |
| Cifar-100-fine (ID) Cifar-10 (OOD) | PSI (our) | $1 - 10^{-13}$ | 0.673 | 0.382 | 0.964 |
| | | (best) $1 - 10^{-7}$ | 0.710 | 0.549 | 0.871 |
| | H&G Baseline [7] | 0.950060 | 0.705 | 0.577 | 0.833 |
| | | (best) $1 - 10^{-1}$ | 0.697 | 0.627 | 0.767 |
| | Energy [10] | 10.93 | 0.702 | 0.552 | 0.853 |

**Virus Dataset – Confusion Matrices and Detailed Analysis of Results**

Figs 7 and 8 show the confusion matrices of the DL classifier with OODs presented as an additional class for two PSI threshold values: $t = 1 - 10^{-1}$ and $t = 1 - 10^{-6}$, respectively. Fig 7 shows that with $t = 1 - 10^{-1}$ some of the OOD samples were erroneously classified as ID samples, mainly as Astrovirus (37), Lassa (99), or Orf (55) viruses. Increasing the PSI threshold to $t = 1 - 10^{-6}$ reduced these numbers but many OODs were still classified as ID samples. In particular, Dengue (5) and Sapovirus (13) samples were misclassified as Astrovirus; Guanarito (22), LCM (4), and Machupo (11) samples were misclassified as Lassa;

and Pseudocowpox (53) samples were misclassified as Orf. The remaining OOD classes: LCM, TBE, and West Nile viruses were much easier to detect as OOD examples. On the other hand, increasing $t$ caused many ID test set samples to be erroneously classified as OODs; over 70% of the test set Ebola and Nipah samples were classified as OOD. Therefore, even though increasing the PSI threshold decreases the number of wrong ID classifications, the high false OOD detection rate makes it a tough compromise that would require a substantial amount of human intervention. Samples detected as OODs should be inspected by a human operator expert that may mark them as new class samples (a novel virus) or add them to the training set (in the case of false OOD detection, which indicates a problematic sample with a feature variant not well represented in the original training set) to improve the representativeness of the set and, thus, to lead to better generalization and performance overall.

|  | | Adenovirus | Astrovirus | CCHF | Cowpox | Ebola | Influenza | Lassa | Marburg | Nipah | Norovirus | Orf | Papilloma | Rift Valley | Rotavirus | OOD | TOTAL |
|---|---|---|---|---|---|---|---|---|---|---|---|---|---|---|---|---|---|
| | Adenovirus | 53 | 0 | 0 | 0 | 0 | 0 | 0 | 0 | 0 | 0 | 0 | 0 | 13 | 0 | 20 | 86 |
| | Astrovirus | 0 | 48 | 0 | 0 | 0 | 0 | 1 | 0 | 0 | 2 | 0 | 0 | 0 | 0 | 15 | 66 |
| | CCHF | 1 | 0 | 71 | 0 | 0 | 0 | 1 | 0 | 0 | 0 | 0 | 0 | 0 | 0 | 13 | 86 |
| | Cowpox | 0 | 0 | 0 | 49 | 0 | 0 | 1 | 0 | 0 | 0 | 0 | 0 | 0 | 0 | 9 | 59 |
| | Ebola | 0 | 0 | 0 | 6 | 230 | 0 | 3 | 12 | 0 | 0 | 0 | 0 | 0 | 0 | 112 | 363 |
| | Influenza | 0 | 0 | 0 | 0 | 0 | 157 | 0 | 0 | 0 | 0 | 0 | 0 | 0 | 0 | 13 | 170 |
| true class | Lassa | 0 | 0 | 0 | 1 | 3 | 0 | 109 | 0 | 0 | 0 | 0 | 0 | 0 | 0 | 15 | 128 |
| | Marburg | 0 | 0 | 0 | 0 | 1 | 2 | 0 | 156 | 0 | 0 | 0 | 0 | 0 | 0 | 14 | 173 |
| | Nipah | 0 | 0 | 0 | 0 | 0 | 1 | 0 | 1 | 27 | 0 | 0 | 0 | 0 | 0 | 6 | 35 |
| | Norovirus | 0 | 0 | 0 | 0 | 0 | 0 | 0 | 0 | 0 | 80 | 0 | 0 | 0 | 0 | 4 | 84 |
| | Orf | 0 | 0 | 0 | 0 | 0 | 0 | 0 | 0 | 0 | 0 | 19 | 0 | 0 | 0 | 12 | 31 |
| | Papilloma | 0 | 0 | 0 | 0 | 0 | 0 | 0 | 0 | 0 | 0 | 0 | 177 | 0 | 0 | 10 | 187 |
| | Rift Valley | 0 | 0 | 2 | 0 | 0 | 0 | 1 | 0 | 2 | 0 | 0 | 0 | 348 | 0 | 39 | 392 |
| | Rotavirus | 0 | 0 | 0 | 0 | 0 | 0 | 0 | 0 | 0 | 0 | 0 | 0 | 0 | 34 | 6 | 40 |
| | OOD | 0 | 37 | 5 | 4 | 1 | 0 | 99 | 1 | 13 | 19 | 55 | 1 | 19 | 5 | 456 | 715 |
| | TOTAL | 54 | 85 | 78 | 60 | 235 | 160 | 215 | 170 | 42 | 101 | 74 | 178 | 380 | 39 | 744 | 2615 |

|  | | Adenovirus | Astrovirus | CCHF | Cowpox | Ebola | Influenza | Lassa | Marburg | Nipah | Norovirus | Orf | Papilloma | Rift Valley | Rotavirus | OOD | TOTAL |
|---|---|---|---|---|---|---|---|---|---|---|---|---|---|---|---|---|---|
| | Dengue | 0 | 16 | 4 | 0 | 0 | 0 | 0 | 0 | 1 | 11 | 0 | 1 | 0 | 2 | 96 | 131 |
| | Guanarito | 0 | 0 | 0 | 1 | 0 | 0 | 44 | 0 | 0 | 0 | 0 | 0 | 0 | 0 | 3 | 48 |
| true OOD class | LCM | 0 | 0 | 0 | 0 | 1 | 0 | 13 | 0 | 1 | 0 | 0 | 0 | 3 | 0 | 37 | 55 |
| | Machupo | 0 | 0 | 0 | 0 | 0 | 0 | 40 | 1 | 6 | 0 | 0 | 0 | 12 | 0 | 49 | 108 |
| | Pseudocowpox | 0 | 0 | 0 | 3 | 0 | 0 | 1 | 0 | 0 | 0 | 55 | 0 | 0 | 0 | 26 | 85 |
| | Sapovirus | 0 | 20 | 0 | 0 | 0 | 0 | 0 | 0 | 0 | 3 | 0 | 0 | 0 | 0 | 15 | 38 |
| | TBE | 0 | 1 | 1 | 0 | 0 | 0 | 0 | 0 | 5 | 5 | 0 | 0 | 0 | 3 | 47 | 62 |
| | WestNile | 0 | 0 | 0 | 0 | 0 | 0 | 1 | 0 | 0 | 0 | 0 | 0 | 4 | 0 | 183 | 188 |
| | TOTAL | 0 | 37 | 5 | 4 | 1 | 0 | 99 | 1 | 13 | 19 | 55 | 1 | 19 | 5 | 456 | 715 |

**Fig 7. Confusion matrix of the TEST and OOD sets with PSI threshold $t = 1 - 10^{-1}$.**

|  | Prediction | | | | | | | | | | | | | | | |
|---|---|---|---|---|---|---|---|---|---|---|---|---|---|---|---|---|
| true class | Adenovirus | Astrovirus | CCHF | Cowpox | Ebola | Influenza | Lassa | Marburg | Nipah | Norovirus | Orf | Papilloma | Rift Valley | Rotavirus | OOD | TOTAL |
| Adenovirus | 34 | 0 | 0 | 0 | 0 | 0 | 0 | 0 | 0 | 0 | 0 | 0 | 6 | 0 | 46 | 86 |
| Astrovirus | 0 | 44 | 0 | 0 | 0 | 0 | 0 | 0 | 0 | 0 | 0 | 0 | 0 | 0 | 22 | 66 |
| CCHF | 0 | 0 | 63 | 0 | 0 | 0 | 1 | 0 | 0 | 0 | 0 | 0 | 0 | 0 | 22 | 86 |
| Cowpox | 0 | 0 | 0 | 47 | 0 | 0 | 0 | 0 | 0 | 0 | 0 | 0 | 0 | 0 | 12 | 59 |
| Ebola | 0 | 0 | 0 | 1 | 90 | 0 | 0 | 5 | 0 | 0 | 0 | 0 | 0 | 0 | 267 | 363 |
| Influenza | 0 | 0 | 0 | 0 | 0 | 117 | 0 | 0 | 0 | 0 | 0 | 0 | 0 | 0 | 53 | 170 |
| Lassa | 0 | 0 | 0 | 1 | 0 | 0 | 64 | 0 | 0 | 0 | 0 | 0 | 0 | 0 | 63 | 128 |
| Marburg | 0 | 0 | 0 | 0 | 0 | 0 | 0 | 88 | 0 | 0 | 0 | 0 | 0 | 0 | 85 | 173 |
| Nipah | 0 | 0 | 0 | 0 | 0 | 0 | 0 | 0 | 8 | 0 | 0 | 0 | 0 | 0 | 27 | 35 |
| Norovirus | 0 | 0 | 0 | 0 | 0 | 0 | 0 | 0 | 0 | 76 | 0 | 0 | 0 | 0 | 8 | 84 |
| Orf | 0 | 0 | 0 | 0 | 0 | 0 | 0 | 0 | 0 | 0 | 24 | 0 | 0 | 0 | 7 | 31 |
| Papilloma | 0 | 0 | 0 | 0 | 0 | 0 | 0 | 0 | 0 | 0 | 0 | 181 | 0 | 0 | 6 | 187 |
| Rift Valley | 0 | 0 | 0 | 0 | 0 | 0 | 0 | 0 | 0 | 0 | 0 | 0 | 262 | 0 | 130 | 392 |
| Rotavirus | 0 | 0 | 0 | 0 | 0 | 0 | 0 | 0 | 0 | 0 | 0 | 0 | 0 | 34 | 6 | 40 |
| OOD | 0 | 18 | 0 | 2 | 0 | 0 | 37 | 0 | 0 | 2 | 53 | 0 | 1 | 0 | 602 | 715 |
| TOTAL | 34 | 62 | 63 | 51 | 90 | 117 | 102 | 93 | 8 | 78 | 77 | 181 | 269 | 34 | 1356 | 2615 |

|  | Prediction | | | | | | | | | | | | | | | |
|---|---|---|---|---|---|---|---|---|---|---|---|---|---|---|---|---|
| true OOD class | Adenovirus | Astrovirus | CCHF | Cowpox | Ebola | Influenza | Lassa | Marburg | Nipah | Norovirus | Orf | Papilloma | Rift Valley | Rotavirus | OOD | TOTAL |
| Dengue | 0 | 5 | 0 | 0 | 0 | 0 | 0 | 0 | 0 | 1 | 0 | 0 | 0 | 0 | 125 | 131 |
| Guanarito | 0 | 0 | 0 | 1 | 0 | 0 | 22 | 0 | 0 | 0 | 0 | 0 | 0 | 0 | 25 | 48 |
| LCM | 0 | 0 | 0 | 0 | 0 | 0 | 4 | 0 | 0 | 0 | 0 | 0 | 0 | 0 | 51 | 55 |
| Machupo | 0 | 0 | 0 | 0 | 0 | 0 | 11 | 0 | 0 | 0 | 0 | 0 | 1 | 0 | 96 | 108 |
| Pseudocowp | 0 | 0 | 0 | 1 | 0 | 0 | 0 | 0 | 0 | 0 | 53 | 0 | 0 | 0 | 31 | 85 |
| Sapovirus | 0 | 13 | 0 | 0 | 0 | 0 | 0 | 0 | 0 | 1 | 0 | 0 | 0 | 0 | 24 | 38 |
| TBE | 0 | 0 | 0 | 0 | 0 | 0 | 0 | 0 | 0 | 0 | 0 | 0 | 0 | 0 | 62 | 62 |
| WestNile | 0 | 0 | 0 | 0 | 0 | 0 | 0 | 0 | 0 | 0 | 0 | 0 | 0 | 0 | 188 | 188 |
| TOTAL | 0 | 18 | 0 | 2 | 0 | 0 | 37 | 0 | 0 | 2 | 53 | 0 | 1 | 0 | 602 | 715 |

**Fig 8. Confusion matrix of the TEST and OOD sets with PSI threshold $t = 1 - 10^{-6}$.**

With the optimal threshold ($t = 0.999828$) 158 of 715 OODs were confused with ID and 516 of 1900 test samples were misclassified as OODs. When we visually inspected these cases about half of them could be explained by the poor image quality. They were out of focus, the virus particles were damaged or decomposed, the debris was touching or partially occluding the target particle, and/or there were image artifacts due to cropping particles close to the image boundary (mirroring). Some of these images were annotated by the pathologists only because the samples were prepared in a way that practically excluded the possibility of cross-contamination, i.e., the experts annotated particles of poor image quality assuming their true class based on the appearance of other virus particles of that specific class and better image quality in that biological sample. This is particularly problematic in images with elongated particles and/or particle clusters. If the image was of inferior quality, all patches cropped from it were likely to be consistently misclassified or detected as OODs. With only a few images for each virus class in the dataset, this constitutes a serious problem that biases the performance metrics.

We also observed that Lassa images have a wide appearance variety in the training set with respect to the particle shape (spherical, elongated, and blobs in between), size, and texture. This resulted in many ambiguous OOD samples being classified as Lassa. Rift Valley is a similar "ambiguous choice" class for relatively small virus particles in images or poor quality due to various decomposition states of the biological sample. Moreover, Guanarito and Lassa were often (33 cases) confused but their images show a lot of similarities in size, shape, and texture. Pseudocowpox and Orf also have a very similar appearance in the images which explains their confusion in so many cases (53). Finally, even though Sapovirus is slightly larger than Astrovirus there are samples very similar to it in the training set which may explain the 15 images being confused.

## Discussion

Softmax assumes that in all classes, the corresponding logit values have the same scale and meaning. However, our investigation of the logit distributions in the training set shows that the decision boundary for the logits is different in each class. As a consequence, the logits from one class cannot be directly compared with the logits from the other class as their distributions have different means and variances. The proposed OOD detector accounts for this inter-class distribution variation. Although there is no clear gain from using our OOD detector over the H&G baseline method in terms of the performance score, our method offers help and guidance in understanding model outputs. It can both aid in interpreting the difficult, borderline cases and facilitate the formation of new classes from the OOD samples. The PSI look-up table distinguishes several types of OODs that can be used as guidance to recognize whether a sample is a combination of two already known classes or yet a completely new specimen for the DL model. Our method may also improve the model performance in a long run by (1) pointing out the ID outliers that should be added to the training set for better data representation and generalization, and (2) interpreting these misclassifications by assigning them to one of the error types from the look-up table.

OOD detection affects the accuracy of the classifier as ID outliers are often detected as OODs. This leads to a decreased number of wrong class classifications, but it also decreases the classification accuracy. An ideal OOD detector would only pick the OOD samples and the misclassified in-distribution samples (which we believe is better than letting them be classified erroneously). However, in real-world applications, it is often very difficult to distinguish an in-distribution outlier and an OOD sample that is close to the training dataset. Changing the threshold of the OOD detector often boils down to a hard compromise; detecting all OOD samples comes at a price of many ID samples being also considered as OOD which substantially increases the manual work. We believe that with a sufficiently varied training set it would be possible to use a very strict OOD detector without overwhelming the experts with false OOD detections. We evaluated our OOD detector on the Virus Dataset acquired with similar instruments, sample preparation, intention, and data handling for both ID and OOD samples. As a consequence, the dataset contains a consistent type of images and related objects. This causes the OOD samples to lay much closer to the ID samples than in comparisons where one dataset is used as ID samples and another – as OOD samples. This is hence representing a real biomedical application and for these reasons much more challenging than the standard benchmarks [21].

The Virus Dataset contains image patches cropped from larger images. In many cases, patches coming from the same original image overlap substantially when the virus particles are clustered or elongated. While extra care was taken to avoid sample leaking between the training, validation, and test sets (they were established at the image level), the overlapping image patches may lead to biased training and evaluation results. During training, for the same number of virus particles in the dataset, the DL model is exposed to less data variation in virus classes that cluster a lot than in those classes that are captured as individual particles and/or small groups of 2-3 particles. During the evaluation, patches from a single "problematic" image (e.g., with an unusual ID sample appearance, or near-distribution OOD samples) with many clustered particles will be consistently misclassified, which results in a much worse performance score than if the classification or OOD detection was performed for the entire image.

The proposed PS function derived from the Naïve Bayes classifier with the normal distribution assumption is only one of the possible methods to compute PS. One could replace the Gaussian model with another distribution model that may better represent the actual data. Another approach would be computing a cumulative distribution function, a logistic function, or other smooth step function and parametrizing it based on the population statistics computed from the data.

In many biomedical applications, large portions of images are discarded and unused. We cropped unused background regions containing various types of debris resembling viruses in their appearance from the training and validation set images, thus, forming the debris set of 736 256 x 256 pixels images. These images can be used as an inexpensive alternative to OOD examples for fine-tuning the OOD detector. We used these images to get the "negative" distribution for each neuron in the last layer of the classifier. These distributions differed from those obtained from the training set. As a consequence, the probability score functions changed, and thus, the thresholds. With this fine-tuning, the OOD detection performance on the virus dataset marginally improved from 0.746 to 0.750. The corresponding threshold was also lower and more robust. The best performance was with $t = 0.75$.

Several methods for data pre-processing to facilitate the separation of ID and OOD sample distributions have been proposed [11, 12]. These methods could be combined with our approach but it would also increase processing time at inference for each sample. Most of these methods require one backpropagation and two forward passes of an image on top of the image modification which may result in too long computation time for some real-world applications. The proposed method requires only a single forward pass and, as it is based on simple calculations and a look-up table, the OOD detection has a constant, negligible processing time. Temperature scaling of logits (as proposed in ODIN [11]) has no impact on our OOD detector performance.

Our results show that, in real-world applications, it is often very difficult to distinguish outliers and unusual data variations from OOD samples. We hope that our study will inspire researchers to focus on realistic datasets and problems for OOD detection.

# Conclusions

The investigation of a deep learning model output is pivotal for understanding its decision processes and assessing its capabilities and limitations. We have analyzed distributions of raw network output vectors (logits before softmax) and observed that each class has its own decision boundary and, thus, the same value has different support for different classes. Inspired by this fact, we developed a new method for out-of-distribution detection. The method is one step beyond simple thresholding of the softmax output towards understanding and interpretation of the model learning process and its output. Our results show that there is space for further development of out-of-distribution detection methods to improve the performance in virus classification and many other realistic and challenging applications.

# Supporting information

S1 File. Distributions of the raw network output for each virus type and the three data subsets: training, validation, and test.

# Interpreting deep learning output for out-of-distribution detection: Supporting information

Damian Matuszewski[1], Ida-Maria Sintorn[1,2]

[1]Department of Information Technology, Uppsala University, Uppsala, Sweden
[2]Vironova AB, Gävlegatan 22, Stockholm, Sweden
damian.matuszewski@it.uu.se, ida.sintorn@it.uu.se

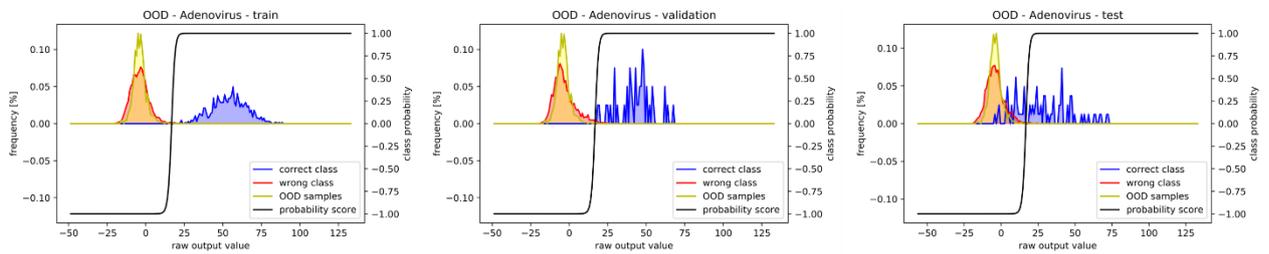

**Figure S 1.** Distributions of the raw network output for the neuron responsible for detecting Adenovirus and the three data subsets: training, validation, and test. The same OOD dataset is overlayed in all three cases.

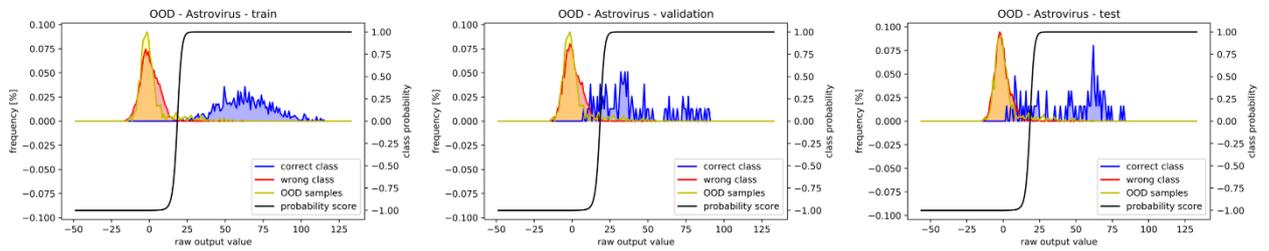

**Figure S 2.** Distributions of the raw network output for the neuron responsible for detecting Astrovirus and the three data subsets: training, validation, and test. The same OOD dataset is overlayed in all three cases.

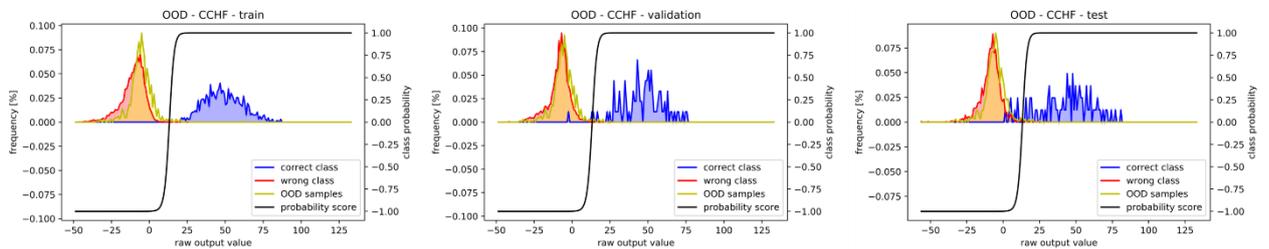

**Figure S 3.** Distributions of the raw network output for the neuron responsible for detecting CCHF and the three data subsets: training, validation, and test. The same OOD dataset is overlayed in all three cases.

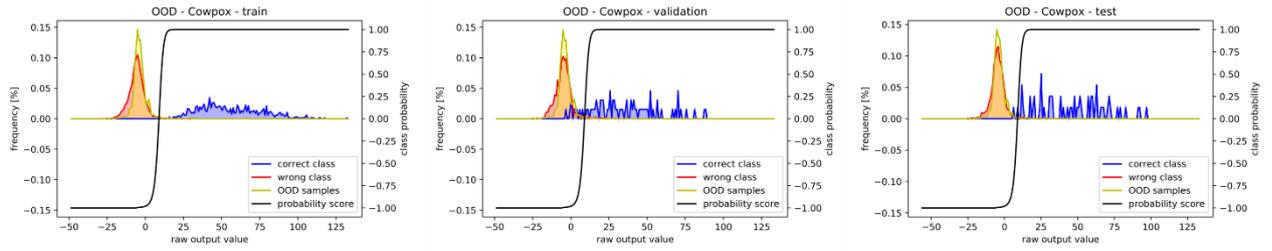

**Figure S 4.** Distributions of the raw network output for the neuron responsible for detecting Cowpox and the three data subsets: training, validation, and test. The same OOD dataset is overlayed in all three cases.

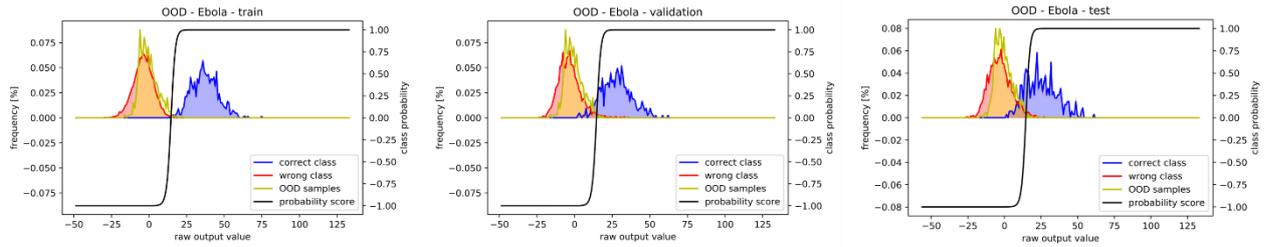

**Figure S 5.** Distributions of the raw network output for the neuron responsible for detecting Ebola and the three data subsets: training, validation, and test. The same OOD dataset is overlayed in all three cases.

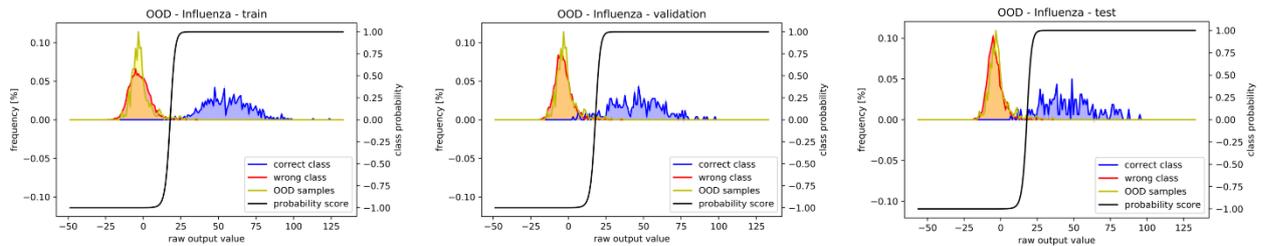

**Figure S 6.** Distributions of the raw network output for the neuron responsible for detecting Influenza and the three data subsets: training, validation, and test. The same OOD dataset is overlayed in all three cases.

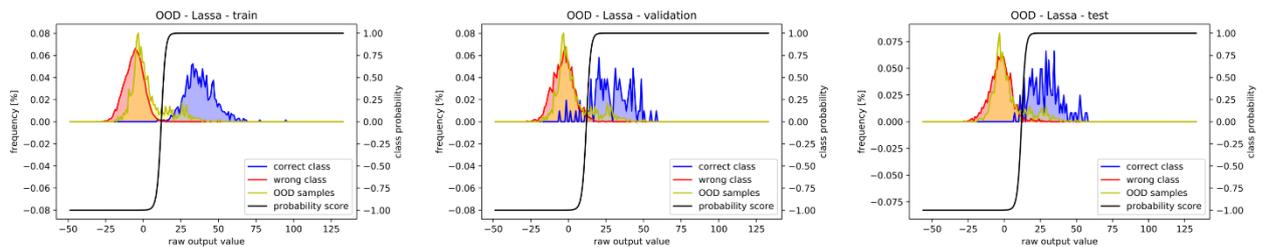

**Figure S 7.** Distributions of the raw network output for the neuron responsible for detecting Lassa and the three data subsets: training, validation, and test. The same OOD dataset is overlayed in all three cases.

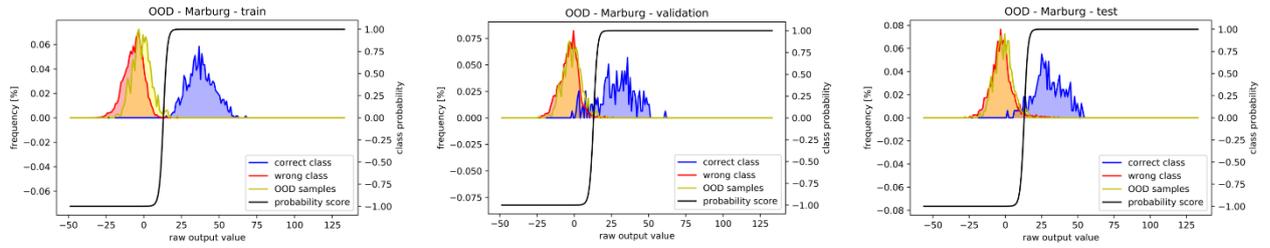

**Figure S 8.** Distributions of the raw network output for the neuron responsible for detecting Marburg and the three data subsets: training, validation, and test. The same OOD dataset is overlayed in all three cases.

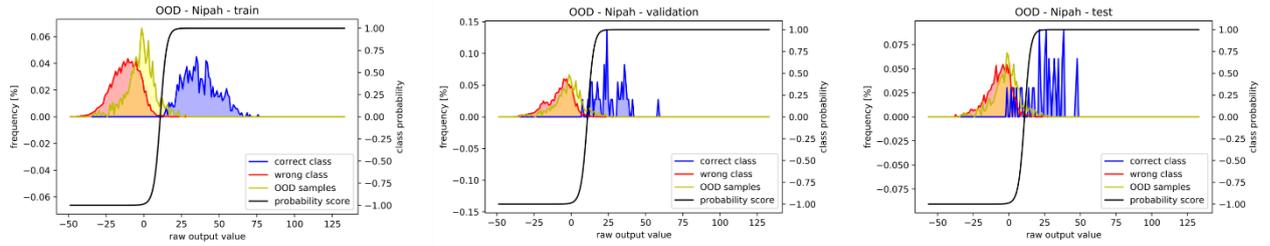

**Figure S 9.** Distributions of the raw network output for the neuron responsible for detecting Nipah and the three data subsets: training, validation, and test. The same OOD dataset is overlayed in all three cases.

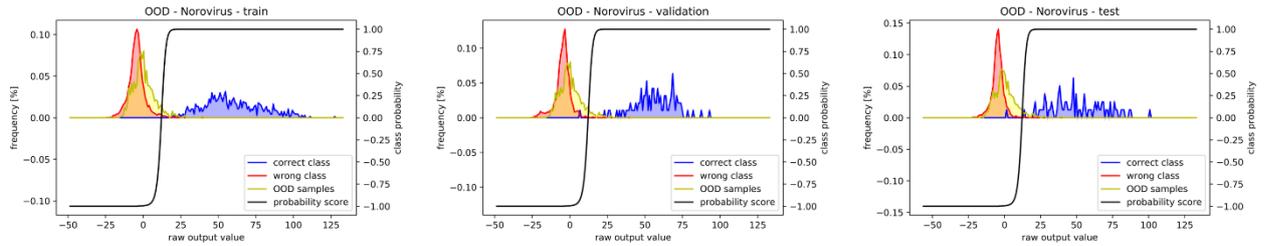

**Figure S 10.** Distributions of the raw network output for the neuron responsible for detecting Norovirus and the three data subsets: training, validation, and test. The same OOD dataset is overlayed in all three cases.

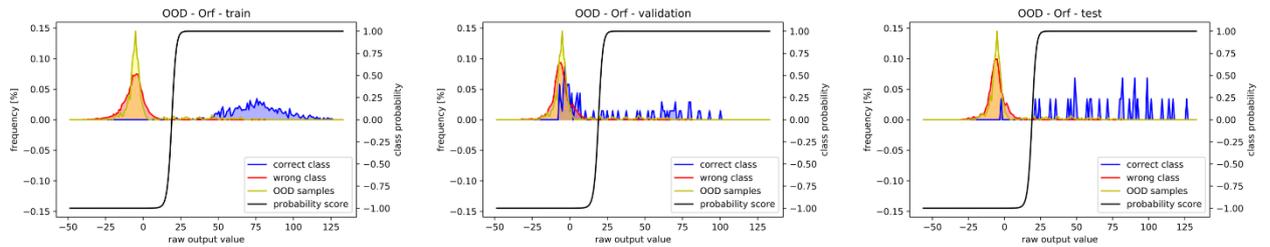

**Figure S 11.** Distributions of the raw network output for the neuron responsible for detecting Orf and the three data subsets: training, validation, and test. The same OOD dataset is overlayed in all three cases.

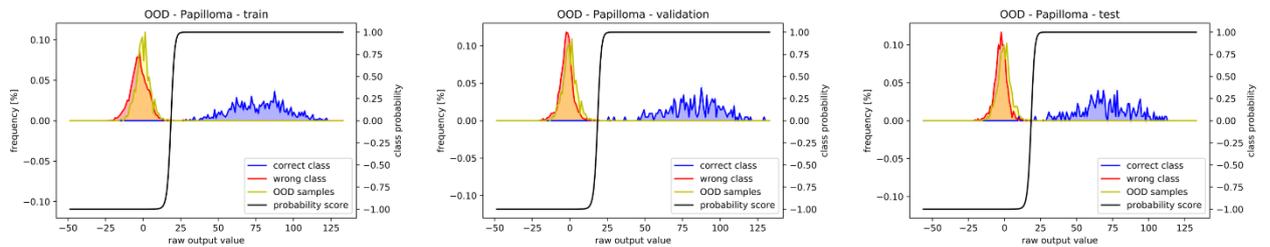

**Figure S 12.** Distributions of the raw network output for the neuron responsible for detecting Papilloma and the three data subsets: training, validation, and test. The same OOD dataset is overlayed in all three cases.

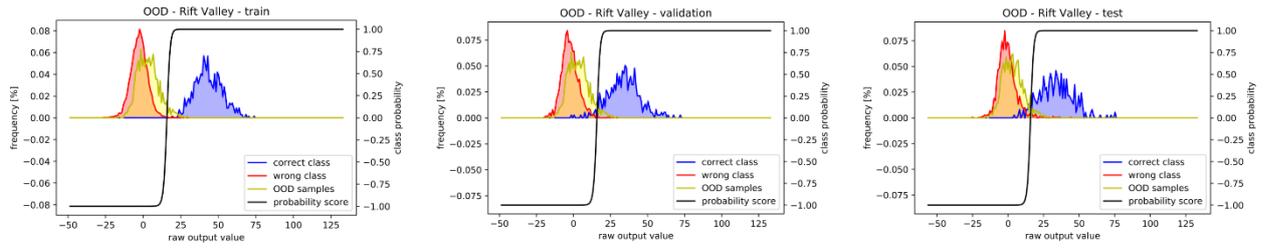

**Figure S 13. Distributions of the raw network output for the neuron responsible for detecting Rift Valley and the three data subsets: training, validation, and test. The same OOD dataset is overlayed in all three cases.**

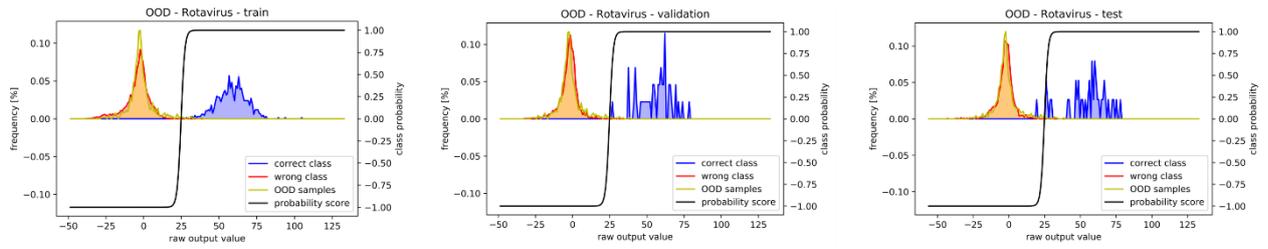

**Figure S 14. Distributions of the raw network output for the neuron responsible for detecting Rotavirus and the three data subsets: training, validation, and test. The same OOD dataset is overlayed in all three cases.**